\documentclass[times,twocolumn,final,authoryear]{elsarticle}

\usepackage{prletters}
\usepackage{framed,multirow}
\usepackage[table,xcdraw]{xcolor}
\usepackage{float}
\floatstyle{plaintop}
\restylefloat{table}
\usepackage[tableposition=top]{caption}

\usepackage{amssymb}
\usepackage{latexsym}
\usepackage{subcaption}

\usepackage{url}
\usepackage{xcolor}
\definecolor{newcolor}{rgb}{.8,.349,.1}

\journal{Pattern Recognition Letters}

\usepackage{graphicx}
\usepackage{bm}
\usepackage{bbm}
\usepackage{amsmath}
\usepackage{amsthm}
\usepackage[ruled,linesnumbered]{algorithm2e}

\usepackage{lineno}

\newtheorem{assumptions1}{Assumption}
\newtheorem{theorem1}{Theorem}

\let\oldnl\nl
\newcommand{\nonl}{\renewcommand{\nl}{\let\nl\oldnl}}

\journal{Journal Name}

\begin{document}

\begin{frontmatter}


\title{Automated Hyperparameter Selection for the PC Algorithm}

\author{Eric V. Strobl}

\address{
Davidson County, Tennessee, United States}

\begin{abstract}
The PC algorithm infers causal relations using conditional independence tests that require a pre-specified Type I $\alpha$ level. PC is however unsupervised, so we cannot tune $\alpha$ using traditional cross-validation. We therefore propose AutoPC, a fast procedure that optimizes $\alpha$ \textit{directly} for a user chosen metric. We in particular force PC to double check its output by executing a second run on the recovered graph. We choose the final output as the one which maximizes stability between the two runs. AutoPC consistently outperforms the state of the art across multiple metrics.
\end{abstract}


\end{frontmatter}
\received{}


\section{Introduction}
Constraint-based causal discovery algorithms recover causal relations from data. In this paper, we focus on PC, the canonical method for inferring causation when no cycles, latent variables or selection bias exist \citep{Spirtes00}. Advances in the PC algorithm naturally lead to progress in more complicated procedures that drop the aforementioned assumptions.

PC recovers causal relations in an unsupervised fashion by executing conditional independence (CI) tests in greedy sequence. As a result, the algorithm requires the user to specify a Type I error rate $\alpha$ a priori. Smaller $\alpha$ values correspond to higher degrees of sparsity. The hyperparameter plays a critical role in determining the accuracy of PC's output in practice. 

Investigators have proposed a few methods for choosing $\alpha$ intelligently. One method assumes a parametric model and chooses the output with the lowest BIC score \citep{Schwarz78}. The score is consistent when the pre-specified likelihood falls within the curved exponential family. The non-parametric StARS algorithm on the other hand runs PC on multiple boostrapped draws and then chooses the $\alpha$ value that maximizes graph stability \citep{Liu10,Raghu18}. The algorithm is computationally intensive and works much better for the graphical lasso than PC. Another method called OCTs chooses $\alpha$ by evaluating the predictive accuracy of Markov boundaries using cross-validation \citep{Biza20}. This algorithm however cannot differentiate between graphs that have equivalent Markov boundaries, so it can prefer structures with too many edges. OCTs also introduces a second user-specified $\alpha$ parameter during cross-validation. Finally, all of the above procedures improve the mean values of some evaluation metrics while maintaining or sacrificing others.

In this paper, we propose to improve the accuracy of PC by optimizing $\alpha$ \textit{directly} for any user chosen metric. We in particular force PC to double check its output according to the metric using a meta-algorithm called AutoPC. We organize the remainder of this paper as follows. We introduce background material in Section \ref{sec_background}. Section \ref{sec_methodology} describes AutoPC in detail. Section \ref{sec_theory} then theoretically justifies AutoPC in both the oracle setting and sample limit. Experiments in Section \ref{sec_experiments} highlight the superiority of AutoPC compared to the aforementioned prior approaches. We conclude the paper in Section \ref{sec_conclusion}.

\section{Background} \label{sec_background}

\subsection{Graphs}

We consider a collection of $d$ random variables $\bm{X}$. A graph contains vertices corresponding to elements of $\bm{X}$. We will use the terms variables and vertices interchangeably. Two vertices are \textit{adjacent} in a graph when they connected by an edge. We use the notation $\textnormal{Adj}(X_i)$ to denote the vertices adjacent to $X_i \in \bm{X}$. We consider directed and undirected edges denoted by $\rightarrow$ and ---, respectively. A \textit{directed graph} only contains directed edges. A \textit{directed path} between $X_i$ and $X_j$ corresponds to a sequence of directed edges between the two vertices. A \textit{collider} refers to the sequence $X_i \rightarrow X_j \leftarrow X_k$ on a directed path. The collider is \textit{unshielded} when $X_i$ and $X_k$ are non-adjacent. A \textit{directed acyclic graph} (DAG) corresponds to a directed graph without \textit{cycles}, or a directed path with edges directed from $X_i$ to $X_j$ and $X_j \rightarrow X_i$. Two vertices $X_i$ and $X_j$ are \textit{d-connected} given $\bm{W} \subseteq \bm{X} \setminus \{X_i,X_j\}$ in a DAG if and only if there exists a directed path between $X_i$ and $X_j$ such that every collider and no non-collider is a member of $\bm{W}$. The vertices are likewise \textit{d-separated} if and only if they are not d-connected.

A density factorizes according to a DAG when it admits the following form:
\begin{equation} \nonumber
    f(\bm{X}) = \prod_{i=1}^d f(X_i | \textnormal{Pa}(X_i)),
\end{equation}
where $\textnormal{Pa}(X_i)$ denotes the \textit{parents} of $X_i$, or variables with directed edges into $X_i$. Such a density satisfies the \textit{global directed Markov property} where $X_i$ and $X_j$ are conditionally independent given $\bm{W}$ if they are d-separated given $\bm{W}$ in the corresponding DAG. We refer to the converse as \textit{d-separation faithfulness}. The \textit{Markov equivalence class} corresponds to the set of DAGs that share the same d-separation and d-connection relations over $\bm{X}$. A \textit{partially directed ayclic graph} (PDAG) contains no cycles and may contain undirected edges. We can summarize the Markov equivalence class using a \textit{completed partially directed acyclic graph} (CPDAG), or a PDAG where any member of the Markov equivalence class corresponds to the PDAG with undirected edges replaced by directed ones.

\subsection{PC Algorithm}
\begin{algorithm}[t]
\SetKwInOut{Input}{Input}
\SetKwInOut{Output}{Output}
\nonl \textbf{Input:} $\alpha$\\
\nonl \textbf{Output:} $\mathbb{G}$\\
 \BlankLine
 
 Form a fully connected undirected graph $\mathbb{G}$\\
 $l \leftarrow -1$ \\
 \Repeat{all pairs of adjacent vertices $(X_i, X_j)$ in $\mathbb{G}$ satisfy $|\textnormal{Adj}(X_i)\setminus X_j| \leq l$}{
 Let $l=l+1$ \\
 \Repeat{all ordered pairs of adjacent vertices $(X_i, X_j)$ in $\mathbb{G}$ with $|\textnormal{Adj}(X_i)\setminus X_j| \geq l$ have been considered}{
 \For{each vertex $X_i$ in $\mathbb{G}$}{
 	Compute $\textnormal{Adj}(X_i)$
 }
 Select a new ordered pair of vertices $(X_i, X_j)$ that are adjacent in $\mathbb{G}$ and satisfy $|\textnormal{Adj}(X_i)\setminus X_j| \geq l$ \\
 
 \Repeat{$X_i$ and $X_j$ are no longer adjacent in $\mathbb{G}$ or all such subsets with $|\bm{W}| = l$ have been considered }{
 
 Choose a new set $\bm{W} \subseteq \textnormal{Adj}(X_i)\setminus X_j$ with $|\bm{W}|=l$ \label{skel:w}\\
 
 Test whether $X_i$ and $X_j$ are independent given $\bm{W}$ to obtain p-value $p$ \label{skel:test}
 
 \If{$p \leq \alpha$}{
 	Delete the edge $X_i - X_j$ from $\mathbb{G}$ \label{skel:delete}
 }

 }
 }
 }

 \caption{Skeleton discovery (stabilized version)} \label{pc_skel}
\end{algorithm}

The PC algorithm tries to infer the CPDAG using CI tests \citep{Spirtes00}. PC first starts with a fully connected undirected graph. The algorithm then proceeds in three phases including skeleton discovery, collider orientation and orientation rule propagation. PC identifies the adjacencies of the CPDAG during skeleton discovery as summarized in Algorithm \ref{pc_skel} \citep{Colombo14}. The algorithm attempts to render any two variables $X_i$ and $X_j$ conditionally independent in lines \ref{skel:w}-\ref{skel:test} using conditioning sets $\bm{W} \subseteq \textnormal{Adj}(X_i) \setminus X_j$ or $\bm{W} \subseteq \textnormal{Adj}(X_j)\setminus X_i$ of increasing size. PC removes the edge $X_i - X_j$ in line \ref{skel:delete} if it succeeds in doing so. Collider orientation identifies unshielded colliders by orienting $X_i \rightarrow X_j \leftarrow X_k$ when $\bm{W}$ renders $X_i$ and $X_k$ conditionally independent with $X_j \not \in \bm{W}$. Finally, orientation rule propagation converts additional undirected edges into directed ones using three graphical criteria proposed in \citep{Meek95}. The final output corresponds to the true CPDAG when d-separation faithfulness holds and the CI test is a \textit{CI oracle} that always makes the correct decision (conditional independence vs. dependence). The CI test is imperfect in practice, so PC yields a PDAG not necessarily equal to the true CPDAG. The CI test also requires the user to pre-specify a Type I error rate $\alpha$. The accuracy of PC, as evaluated by any of the metrics described below, depends heavily on the choice of $\alpha$. 

\subsection{Metrics}
Many different methods exist for evaluating the output of PC. The most popular metric is the structural Hamming distance (SHD) which corresponds to the number of insertions, deletions or flips to transform the PDAG output into the true CPDAG \citep{Tsamardinos06}. Investigators also frequently use the F1 score and Matthew's correlation coefficient (MCC) to compare against the true CPDAG, where positives correspond to the presence of edges and negatives to their absence. The structural intervention distance (SID) counts the number of interventional distributions that can differ as compared to the true DAG \citep{Peters15}. The authors suggested a strategy for computing SID against the true CPDAG as well, but they did not implement or evaluate a practical approach.

\section{Methodology} \label{sec_methodology}

\begin{algorithm}[t]
 \nonl \textbf{Input:} Ordered Type I error rates $\mathcal{A}$\\
 \nonl \textbf{Output:} PDAG $\mathbb{G}^*$\\
 \BlankLine

$\eta \leftarrow -\infty$\\
\For{each $\alpha \in \mathcal{A}$}{
Run PC with $\alpha$ to obtain $\mathbb{G}_\alpha^1$ \label{AutoPC:PC1}

Run PC again with $\alpha$ but with line \ref{skel:w} of Algorithm \ref{pc_skel} modified with $\bm{W} \subseteq [\textnormal{Adj}(X_i) \cap \textnormal{Pa}_{\mathbb{G}_\alpha^1}(X_i)] \setminus X_j$ to obtain $\mathbb{G}_\alpha^2$ \label{AutoPC:PC2}

\If{$m(\mathbb{G}_\alpha^1, \mathbb{G}_\alpha^2) > \eta$ \label{AutoPC:compare}}{
$\mathbb{G}^* \leftarrow \mathbb{G}_\alpha^1$

$\eta \leftarrow m(\mathbb{G}_\alpha^1, \mathbb{G}_\alpha^2)$

\textbf{break} if $\eta = \textnormal{perfect score}$ \label{AutoPC:break}
}
}

 \caption{AutoPC} \label{alg_AutoPC}
\end{algorithm}

We now introduce AutoPC in Algorithm \ref{alg_AutoPC}. Let $\mathcal{A}$ denote the set of possible $\alpha$ values ordered from lowest to highest. For each $\alpha \in \mathcal{A}$, AutoPC first runs the usual PC algorithm in line \ref{AutoPC:PC1} in order to recover the PDAG $\mathbb{G}_\alpha^1$. AutoPC then runs the PC algorithm again in line \ref{AutoPC:PC2} using $\alpha$ but with $\bm{W} \subseteq \textnormal{Adj}(X_i) \setminus X_j$ in line \ref{skel:w} of Algorithm \ref{pc_skel} replaced with $\bm{W} \subseteq [\textnormal{Adj}(X_i) \cap \textnormal{Pa}_{\mathbb{G}_\alpha^1}(X_i)] \setminus X_j$; the notation $\textnormal{Pa}_{\mathbb{G}_\alpha^1}(X_i)$ refers to vertices with directed edges into $X_i$ as well as those sharing an undirected edge with $X_i$ in $\mathbb{G}_\alpha^1$. AutoPC therefore obtains two CPDAG estimates $\mathbb{G}^1_\alpha$ and $\mathbb{G}^2_\alpha$. The procedure compares the two in line \ref{AutoPC:compare} using a bounded user chosen metric $m$, which we seek to maximize. The algorithm finally outputs the PDAG associated with the $\alpha$ value that maximizes $m(\mathbb{G}^1_\alpha, \mathbb{G}^2_\alpha)$.

AutoPC thus runs the PC algorithm twice on each $\alpha \in \mathcal{A}$. The second run is however much faster because the conditioning sets are restricted to subsets of the parents from the previous PDAG. The second run also serves to check the first according to the user chosen metric $m$.

\section{Theory} \label{sec_theory}

\subsection{Oracle Property}
We now justify AutoPC in the oracle setting. We assume access to a bounded metric $m$ that can compare two PDAGs. Without loss of generality, let the metric take values on the interval $[0,1]$ with $m(\mathbb{G}, \mathbb{G})=1$, where $\mathbb{G}$ denotes the true CPDAG. We therefore seek to maximize $m$. We can appropriately invert or normalize the SHD, MCC and other evaluation criteria to meet these requirements.
 
We first show that AutoPC run with a CI oracle always returns $m(\mathbb{G}^1_{\alpha}, \mathbb{G}^2_{\alpha}) = 1$ for any $0<\alpha<1$. We of course have $\mathbb{G}^1_{\alpha} = \mathbb{G}$ for the usual PC algorithm. The following statement also holds:
\begin{theorem1}
We have $\mathbb{G}^2_{\alpha} = \mathbb{G}$ under d-separation faithfulness.
\end{theorem1}
\begin{proof}
Note that $X_i$ and $X_j$ are d-separated given some $\bm{W} \subseteq \bm{X} \setminus \{X_i,X_j\}$ if and only if $X_i$ and $X_j$ are d-separated given some $\bm{W}$ where $\bm{W} \subseteq \textnormal{Pa}(X_i) \setminus X_j$ or $\bm{W} \subseteq \textnormal{Pa}(X_j) \setminus X_i$ \citep{Spirtes00}. Thus $\mathbb{G}^2_{\alpha}$ has the same skeleton as $\mathbb{G}^1_{\alpha}$. Moreover, we have an unshielded collider $X_i \rightarrow X_k \leftarrow X_j$ in $\mathbb{G}^1_{\alpha}$ if and only if the same structure is present in $\mathbb{G}^2_{\alpha}$ for two reasons. For the forward direction, if $X_i \rightarrow X_j \leftarrow X_k$ in $\mathbb{G}^1_{\alpha}$, then $X_j \not \in \bm{W}$ in the second run by construction of AutoPC. For the backward direction, if the unshielded collider exists in $\mathbb{G}^2_{\alpha}$, then there exists a d-separating set $\bm{W}$ that does not contain $X_j$ for $\mathbb{G}^1_{\alpha}$. The conclusion follows because the Markov equivalence class contains all graphs with the same skeleton and unshielded v-structures (Corollary 3.2 in \citep{Verma92}).
\end{proof}
\noindent As a result, the second run of PC still returns $\mathbb{G}$ even though the algorithm reruns the PC algorithm on restricted conditioning sets. This implies that $m$ is maximized to one. 

Intuitively then, running AutoPC with CI tests should converge to the oracle setting. We in particular run PC with a sequence $\alpha_n$ and then again on the recovered graph with the same $\alpha_n$, where $n$ denotes sample size. These two steps should discover the exact same graph in the infinite sample limit if we choose $\alpha_n$ appropriately such that PC itself is consistent. As a result, $m(\mathbb{G}^1_{\alpha_n}, \mathbb{G}^2_{\alpha_n})$ should converge to one in probability even if we replace the CI oracle with a CI test.

\subsection{Consistency}
We confirm the above intuition rigorously as well as derive an even better result; AutoPC automatically discovers an appropriate sequence $\widehat{\alpha}_n$ that maximizes $m$. The argument is inspired by \citep{Sun13}. We write $m(\mathbb{G}^1_{\alpha}, \mathbb{G}^2_{\alpha})$ as $m(\alpha)$ to simplify notation. Consider a sequence $\widehat{\alpha}_n \in \{ \alpha : m(\alpha) > 1 - \gamma_n\} = \mathcal{A}_n$ with strictly positive $\gamma_n \rightarrow 0$. In other words, the hyperparameter set $\mathcal{A}_n$ contains $\alpha$ values that are ``good enough'' to maximize $m$.

We consider the following assumption:
\begin{assumptions1} \label{assump_hyper}
There exists a sequence of hyperparameters $\alpha_n$ such that PC is CPDAG consistent, i.e. $\mathbb{P}(\mathbb{G}_{\alpha_n} = \mathbb{G}) \geq 1 - \varepsilon_n$ for some $\varepsilon_n \rightarrow 0$.
\end{assumptions1}
\noindent The above assumption is for example already known to hold for the PC algorithm in the Gaussian case by setting $\alpha_n = 2(1 - \Phi(\sqrt{n} c /2))$, where $c$ denotes the (fixed) minimum conditional correlation coefficient in Fisher's z-test when conditional dependence holds \citep{Kalisch07}. Analyzing the proof structure in that paper, we can see that a similar statement holds in the more general case so long as we can lower bound the estimand of the CI test statistic.

We have the following consistency result:
\begin{theorem1} \label{thm_metric}
(Metric consistency) If Assumption \ref{assump_hyper} holds, then $\mathbb{P}(m(\widehat{\alpha}_n) > 1 - 2 \gamma_n) \rightarrow 1$ for any $\gamma_n$ converging to zero slower than $\varepsilon_n$.
\end{theorem1}
\begin{proof}
We can write $m(\widehat{\alpha}_n)> 1-\gamma_n \geq m(\alpha_n)(1-\gamma_n)$. We therefore have:
\begin{equation}
\begin{aligned}
    \mathbb{P}(m(\widehat{\alpha}_n) > 1 - 2 \gamma_n) &\geq \mathbb{P}(m(\widehat{\alpha}_n) > (1-\gamma_n)^2)\\
    &\geq \mathbb{P}(m(\alpha_n) > 1-\gamma_n). \label{eq:b1}
\end{aligned}
\end{equation}
We can upper bound one minus the term at the bottom by the Markov inequality:
\begin{equation}\nonumber
\mathbb{P}(1 - m(\alpha_n) \geq \gamma_n) \leq \frac{\mathbb{E}(1 - m(\alpha_n))}{\gamma_n}.
\end{equation}
We have the following sequence for $\mathbb{E}(m(\alpha_n))$:
\begin{equation}\nonumber
\mathbb{E}(m(\alpha_n)) \geq \mathbb{E}(\mathbbm{1}_{m(\alpha_n) = 1}) = \mathbb{P} ( m(\alpha_n) = 1 ).
\end{equation}
We can lower bound the later term:
\begin{equation}\nonumber
\begin{aligned}
\mathbb{P} ( m(\alpha_n) = 1 ) &\geq \mathbb{P}(\mathbb{G}^1_{\alpha_n} = \mathbb{G}, \mathbb{G}^2_{\alpha_n} = \mathbb{G})\\
&= \mathbb{P}(\mathbb{G}^2_{\alpha_n} = \mathbb{G}| \mathbb{G}^1_{\alpha_n} = \mathbb{G})  \mathbb{P}(\mathbb{G}^1_{\alpha_n} = \mathbb{G})\\
&\geq (1-\varepsilon_n)^2 \geq 1-2\varepsilon_n.
\end{aligned}
\end{equation}
Returning back then to Equation \eqref{eq:b1}, we have:
\begin{equation} \nonumber
    \mathbb{P}(m(\widehat{\alpha}_n) > 1-2\gamma_n) \geq 1 - \frac{2\varepsilon_n}{ \gamma_n}.
\end{equation}
The conclusion follows because $\gamma_n$ converges to zero slower than $\varepsilon_n$.
\end{proof}
\noindent In other words, $m(\widehat{\alpha}_n)$ probabilistically approaches the oracle value one. This implies that AutoPC discovers a sequence $\widehat{\alpha}_n$ that maximizes $m$, so long as $\mathcal{A}_n$ has entries greater than $1-\gamma_n$.

We of course want to eliminate edge cases such as $\mathcal{A}_n = \{0\}$ with $m=1$ always because these trivially maximize $m$ without helping PC converge to $\mathbb{G}$. We therefore also consider the following assumption with the set of PDAGs $\mathcal{G}$:
\begin{assumptions1} \label{assump_metric}
We have $\mathbb{G}^1_{\alpha_n} \in \mathcal{G}$ and $m(\alpha_n)=1$ only if $\mathbb{G}_{\alpha_n}^1 = \mathbb{G}$ for any $\alpha_n \in \mathcal{A}_n $ with $n$ large enough.
\end{assumptions1}
\noindent Stated differently, $\mathcal{A}_n$ restricts PC to $\mathcal{G}$, but only $\mathbb{G} \in \mathcal{G}$ maximizes the metric. This is a much weaker requirement than finding a sequence of $\alpha$ values such that PC can achieve CPDAG consistency; $\mathcal{A}_n$ only needs to confine the solution to a ballpark where $m$ can sort out the rest. The assumption is also reasonable with common metric choices such as the SHD, F1 and MCC because all of these return one when $\mathbb{G}^1_{\alpha_n} = \mathbb{G}^2_{\alpha_n}$. Formally:
\begin{theorem1}
(CPDAG consistency) If Assumptions \ref{assump_hyper} and \ref{assump_metric} hold, then $\mathbb{P}(\mathbb{G}^1_{\widehat{\alpha}_n} = \mathbb{G}) \rightarrow 1$ for any $\gamma_n$ converging to zero slower than $\varepsilon_n$.
\end{theorem1}
\begin{proof}
We can lower bound the quantity $\mathbb{P}(\mathbb{G}^1_{\widehat{\alpha}_n}= \mathbb{G})$ for $n$ large enough by the following in light of Assumption \ref{assump_metric}:
\begin{equation} \label{eq_thm3}
\begin{aligned}
    &\mathbb{P}(m(\alpha_n) > 1 - \gamma_n, \mathbb{G}^1_{\alpha_n} \in \mathcal{G})\\
    \geq \hspace{1mm}& \mathbb{P}(m(\alpha_n) > 1 - \gamma_n, \mathbb{G}^1_{\alpha_n}=\mathbb{G})\\
    = \hspace{1mm}&\mathbb{P}(m(\alpha_n) > 1 - \gamma_n 
    |\mathbb{G}^1_{\alpha_n}=\mathbb{G}) \mathbb{P}(\mathbb{G}^1_{\alpha_n} = \mathbb{G}).
\end{aligned}
\end{equation}
Bounding one minus the first term by the Markov inequality and then proceeding as in Theorem \ref{thm_metric}, we obtain 
\begin{equation} \nonumber
\begin{aligned}
    \mathbb{P}(m(\alpha_n) = 1 | \mathbb{G}^1_{\alpha_n}=\mathbb{G}) &\geq \mathbb{P}(\mathbb{G}^2_{\alpha_n} =\mathbb{G} | \mathbb{G}^1_{\alpha_n}=\mathbb{G})\\
    &\geq 1 - \varepsilon_n.
\end{aligned}
\end{equation}
Hence, the bottom of Equation \eqref{eq_thm3} is bounded below:
\begin{equation} \nonumber
\begin{aligned}
    &\mathbb{P}(m(\alpha_n) > 1 - \gamma_n 
    |\mathbb{G}^1_{\alpha_n}=\mathbb{G}) \mathbb{P}(\mathbb{G}^1_{\alpha_n} = \mathbb{G})\\
    \geq \hspace{1mm}& \Big(1-\frac{\varepsilon_n}{\gamma_n} \Big) (1-\varepsilon_n)\\
    \geq \hspace{1mm}& 1 - \varepsilon_n -\frac{\varepsilon_n}{\gamma_n}
\end{aligned}
\end{equation}
The conclusion follows because $\gamma_n$ converges to zero slower than $\varepsilon_n$.
\end{proof}

In practice, $\mathcal{A}_n$ is finite with a fixed sample size. The set must contain reasonable choices of $\alpha$ that help maximize an informative $m$. We cannot expect AutoPC to perform well if all of the hyperparameter choices are lousy or the metric does not differentiate between the PDAGs in the first place. These pathological situations are therefore excluded by Assumptions \ref{assump_hyper} and \ref{assump_metric}, respectively.

\subsection{Complexity}
PC is already known to run in $O(d^{q})$ time, where $d$ denotes the number of vertices and $q$ the maximum number of vertices adjacent to any vertex \citep{Spirtes00}. The second run of PC in Algorithm \ref{alg_AutoPC} however only takes $O(d^r)$ time, where $r$ denotes the maximum number of \textit{parents} and vertices adjacent with an undirected edge. This step therefore completes much faster than the first. We can predict that AutoPC will only take slightly longer than PC run on all $\alpha$ values in $\mathcal{A}$.

\section{Experiments} \label{sec_experiments}
\subsection{Setup}
We instantiated 500 Gaussian DAGs with $d=10$, 20 or 50 variables, an expected neighborhood size of 2 and coefficients drawn uniformly from $[-1.3,-0.3] \cup [0.3,1.3]$. We drew sample sizes from $\{1000,10000,100000\}$ creating a total of $500 \times 3 \times 3 = 4500$ datasets. We then ran BIC, StARS, OCTs and AutoPC with default parameters on each dataset using PC equipped with Fisher's z-test as well as $\alpha$ choices $\{0.0005, 0.001, 0.005, 0.01, 0.05, 0.1\}=\mathcal{A}$. We evaluated the algorithms using SHD, F1 and MCC that all allow us to compare the recovered PDAG against the true CPDAG; we cannot apply SID as mentioned previously. We also recorded time.

\subsection{Synthetic Data}

We summarize the accuracy results in Tables 1 (a) to (c); ``Mean'' denotes mean performance across the $\alpha$ values. AutoPC obtained the best average scores across all three accuracy criteria without any exceptions. All pair-wise comparisons at each sample size and dimension were significant at a Bonferonni corrected threshold of $0.05/4$ using paired t-tests. We conclude that AutoPC easily sets the new state of the art for hyperparameter selection in PC. 

Table 1 (d) also lists the timing results in seconds. Some values are negative because AutoPC breaks early if it achieves a perfect score in line \ref{AutoPC:break}. AutoPC rivals BIC while outperforming OCTs and StARS by around an order of magnitude. We conclude that AutoPC is also fast.

\begin{figure}[t]
\begin{subfigure}{0.225\textwidth}
    \centering
    \includegraphics[scale=0.425]{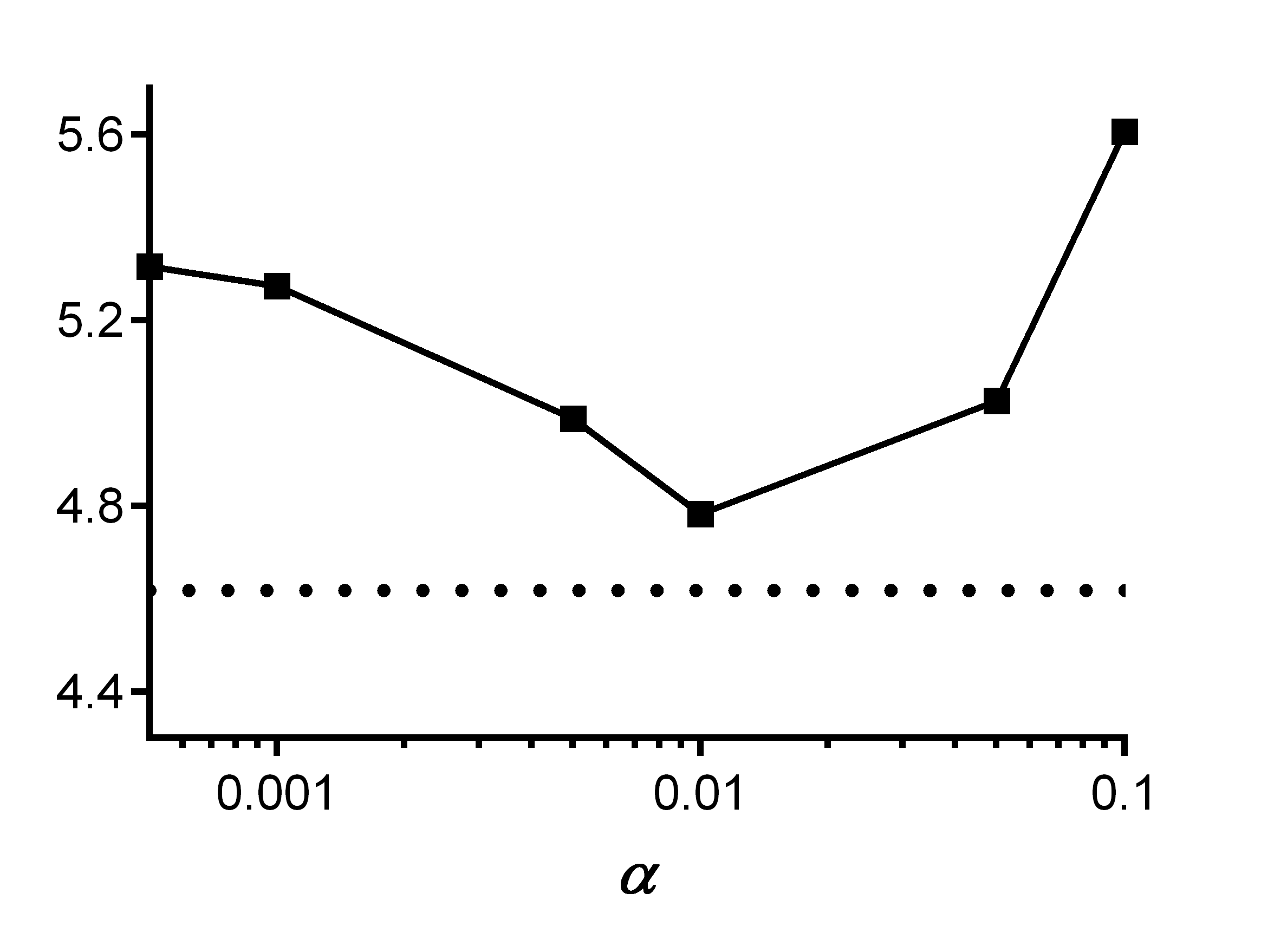}
    \caption{SHD}
\end{subfigure}
\begin{subfigure}{0.225\textwidth}
    \centering
    \includegraphics[scale=0.425]{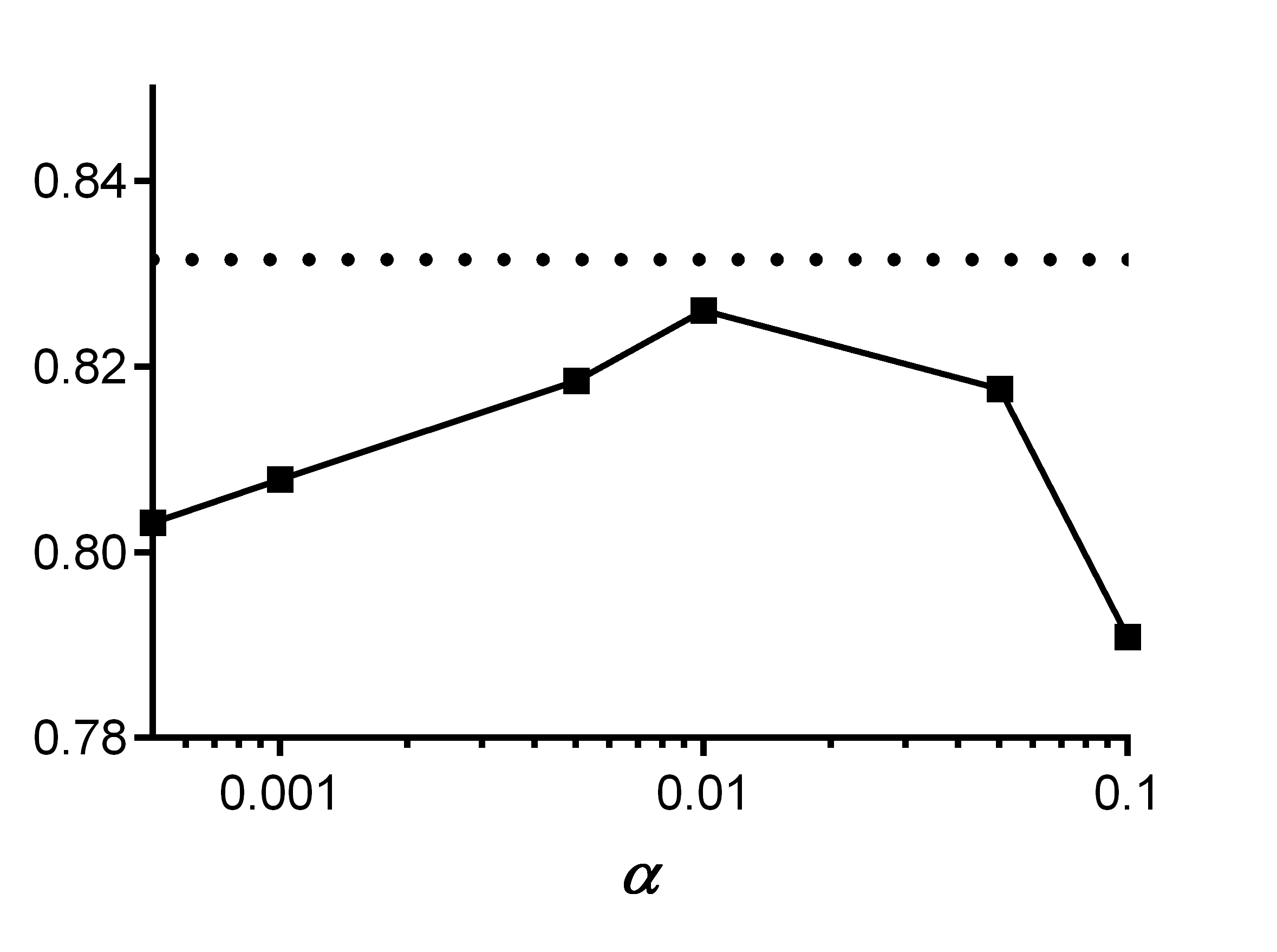}
    \caption{F1}
\end{subfigure}

\begin{subfigure}{0.225\textwidth}
    \centering
    \includegraphics[scale=0.425]{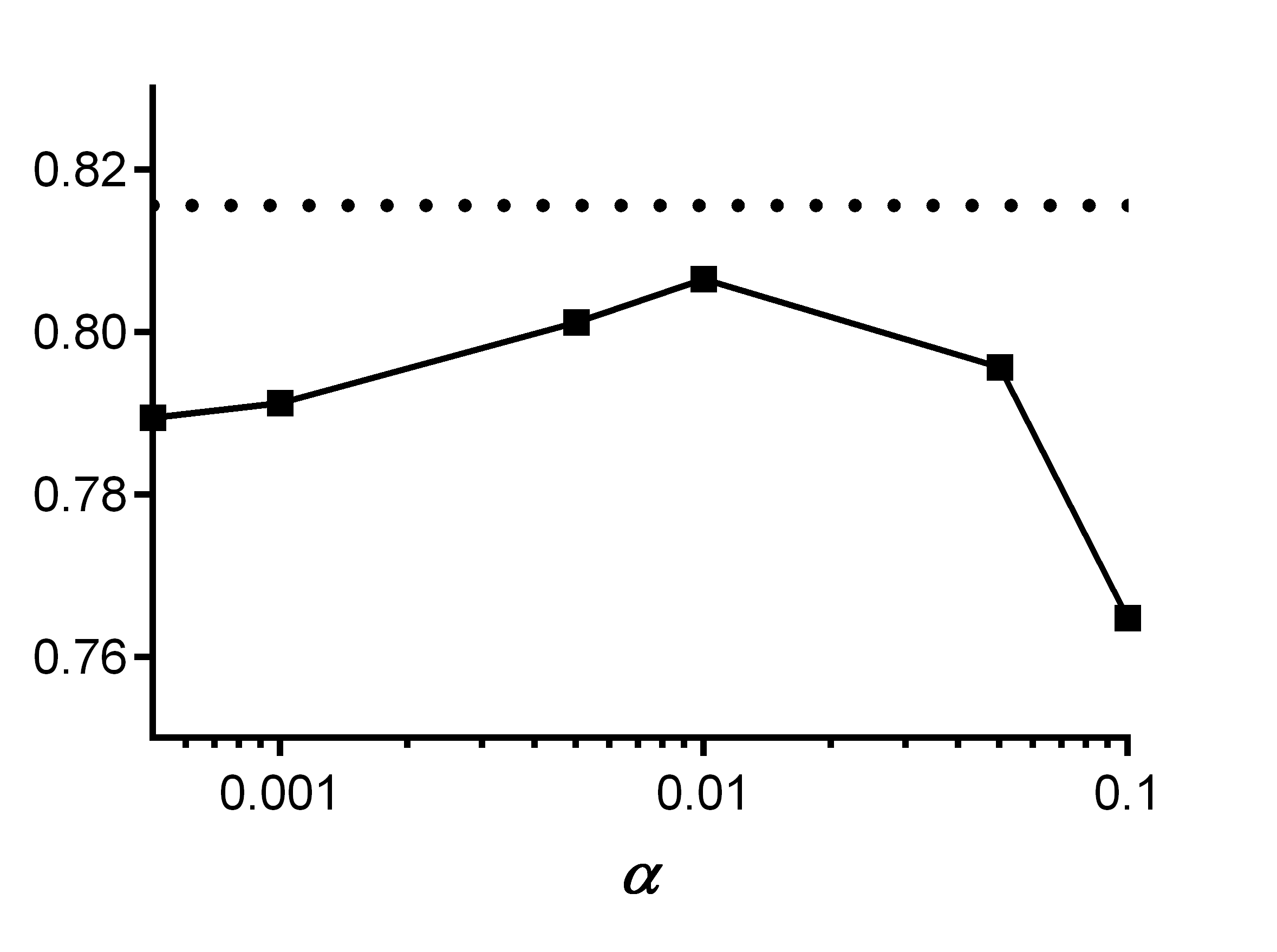}
    \caption{MCC}
\end{subfigure}
\caption{Performance of AutoPC versus PC across $\alpha$ values for $n=1000, d=10$. The dotted line corresponds to AutoPC and the solid line to the average accuracy of PC run at different $\alpha$ values. AutoPC outperforms PC run at any $\alpha$ value on average.} \label{fig_alphas}
\end{figure}

We compare AutoPC against the average PC run with each $\alpha$ value in Figure \ref{fig_alphas}. AutoPC even consistently outperformed PC equipped with the best average $\alpha$ value across all metrics; the graphs for the other sample sizes and dimensions look similar. The proposed algorithm therefore customizes $\alpha$ according to each graphical structure. 

\subsection{Real Data}
We next ran the algorithms on data collected from the Framingham Heart Study \citep{Mahmood14}. The dataset contains repeated measurements of eight variables over three time steps as well as 2003 samples after removing instances with missing values. We do not know the entire causal graph in this case, but we do know a partial ground truth; variables in future time steps cannot cause past ones, and any variable in one time step must directly cause its corresponding variable in the next time step. We ran the algorithms on $500$ bootstrapped draws using $\mathcal{A} = \{$5E-8,  1E-7, 5E-7, 1E-5, 5E-6, 1E-5$\}$; the causal relations are strong here, so the algorithms just choose the smallest $\alpha$ value if we use the original set in the previous subsection.

We summarize results in Table \ref{table_real} again with SHD, F1, MCC and time. The results replicate those seen in the synthetic data; AutoPC dominates the accuracy criteria. The pairwise comparisons are again all significant at a Bonferroni corrected threshold of 0.05/4. The algorithm also comes in at second place behind BIC in terms of computation time.

\begin{table}[b!]
\centering
\begin{subtable}{0.45\textwidth}
  \centering
\smaller
\begin{tabular}{
>{\columncolor[HTML]{FFFFFF}}c 
>{\columncolor[HTML]{FFFFFF}}c |ccccc}
\cellcolor[HTML]{EFEFEF}\textit{n} & \cellcolor[HTML]{EFEFEF}\textit{d} & \cellcolor[HTML]{EFEFEF}Mean & \cellcolor[HTML]{EFEFEF}BIC & \cellcolor[HTML]{EFEFEF}StARS & \cellcolor[HTML]{EFEFEF}OCTs & \cellcolor[HTML]{EFEFEF}AutoPC \\ \hline
1,000                               & 10                                 & 5.165                          & 4.732                       & 5.316                         & 4.966                        & \textbf{4.618}                 \\
10,000                             & 10                                 & 3.483                          & 3.008                       & 3.438                         & 3.134                        & \textbf{2.832}                 \\
100,000                            & 10                                 & 2.145                          & 1.766                       & 1.840                         & 1.896                        & \textbf{1.510}                 \\ \hline
1,000                               & 20                                 & 9.603                          & 9.000                       & 9.446                         & 9.394                        & \textbf{8.560}                 \\
10,000                             & 20                                 & 6.276                          & 5.416                       & 5.244                         & 5.780                        & \textbf{4.700}                 \\
100,000                            & 20                                 & 4.554                          & 3.728                       & 3.346                         & 3.620                        & \textbf{2.908}                 \\ \hline
1,000                               & 50                                 & 24.36                         & 21.03                      & 27.46                        & 24.69                       & \textbf{18.33}                \\
10,000                             & 50                                 & 16.50                         & 11.60                      & 15.83                        & 10.22                       & \textbf{8.500}                 \\
100,000                            & 50                                 & 13.47                         & 8.400                       & 10.54                        & 5.992                        & \textbf{4.912}     
\end{tabular}
\caption{SHD} \label{fig_SHD}
\end{subtable}

\vspace{5mm}
\begin{subtable}{0.45\textwidth}
  \centering
\smaller
\begin{tabular}{
>{\columncolor[HTML]{FFFFFF}}c 
>{\columncolor[HTML]{FFFFFF}}c |ccccc}
\cellcolor[HTML]{EFEFEF}\textit{n} & \cellcolor[HTML]{EFEFEF}\textit{d} & \cellcolor[HTML]{EFEFEF}Mean & \cellcolor[HTML]{EFEFEF}BIC & \cellcolor[HTML]{EFEFEF}StARS & \cellcolor[HTML]{EFEFEF}OCTs & \cellcolor[HTML]{EFEFEF}AutoPC \\ \hline
1,000                              & 10                                 & 0.810                          & 0.823                       & 0.803                         & 0.827                        & \textbf{0.831}                 \\
10,000                             & 10                                 & 0.890                          & 0.912                       & 0.896                         & 0.905                        & \textbf{0.916}                 \\
100,000                            & 10                                 & 0.943                          & 0.955                       & 0.952                         & 0.957                        & \textbf{0.964}                 \\ \hline
1,000                              & 20                                 & 0.810                          & 0.824                       & 0.812                         & 0.816                        & \textbf{0.829}                 \\
10,000                             & 20                                 & 0.878                          & 0.896                       & 0.897                         & 0.892                        & \textbf{0.905}                 \\
100,000                            & 20                                 & 0.920                          & 0.938                       & 0.947                         & 0.940                        & \textbf{0.952}                 \\ \hline
1,000                              & 50                                 & 0.804                          & 0.824                       & 0.786                         & 0.802                        & \textbf{0.845}                 \\
10,000                             & 50                                 & 0.872                          & 0.906                       & 0.871                         & 0.922                        & \textbf{0.928}                 \\
100,000                            & 50                                 & 0.894                          & 0.943                       & 0.908                         & 0.953                        & \textbf{0.960}                
\end{tabular}
\caption{F1} \label{fig_F1}
\end{subtable}

\vspace{5mm}
\begin{subtable}{0.45\textwidth}
\centering
\smaller
\begin{tabular}{
>{\columncolor[HTML]{FFFFFF}}c 
>{\columncolor[HTML]{FFFFFF}}c |ccccc}
\cellcolor[HTML]{EFEFEF}\textit{n} & \cellcolor[HTML]{EFEFEF}\textit{d} & \cellcolor[HTML]{EFEFEF}Mean & \cellcolor[HTML]{EFEFEF}BIC & \cellcolor[HTML]{EFEFEF}StARS & \cellcolor[HTML]{EFEFEF}OCTs & \cellcolor[HTML]{EFEFEF}AutoPC \\ \hline
1,000                              & 10                                 & 0.791                          & 0.809                       & 0.789                         & 0.803                        & \textbf{0.815}                 \\
10,000                             & 10                                 & 0.874                          & 0.893                       & 0.876                         & 0.890                        & \textbf{0.900}                 \\
100,000                            & 10                                 & 0.931                          & 0.947                       & 0.942                         & 0.949                        & \textbf{0.958}                 \\ \hline
1,000                              & 20                                 & 0.800                          & 0.817                       & 0.804                         & 0.806                        & \textbf{0.822}                          \\
10,000                             & 20                                 & 0.879                          & 0.902                       & 0.900                         & 0.895                        & \textbf{0.911}                 \\
100,000                            & 20                                 & 0.912                          & 0.928                       & 0.939                         & 0.929                        & \textbf{0.944}                 \\ \hline
1,000                              & 50                                 & 0.805                          & 0.830                       & 0.781                         & 0.806                        & \textbf{0.852}                 \\
10,000                             & 50                                 & 0.869                          & 0.905                       & 0.870                         & 0.915                        & \textbf{0.927}                 \\
100,000                            & 50                                 & 0.894                          & 0.941                       & 0.911                         & 0.955                        & \textbf{0.958}                
\end{tabular}
\caption{MCC} \label{fig_MCC}
\end{subtable}

\vspace{5mm}
\begin{subtable}{0.45\textwidth}
\centering
\smaller
\begin{tabular}{
>{\columncolor[HTML]{FFFFFF}}c 
>{\columncolor[HTML]{FFFFFF}}c |cccc}
\cellcolor[HTML]{EFEFEF}\textit{n} & \cellcolor[HTML]{EFEFEF}\textit{d} & \cellcolor[HTML]{EFEFEF}BIC & \cellcolor[HTML]{EFEFEF}StARS & \cellcolor[HTML]{EFEFEF}OCTs & \cellcolor[HTML]{EFEFEF}AutoPC \\ \hline
1,000                              & 10                                 & 0.048                       & 0.500                         & 1.842                        & \textbf{0.021}                 \\
10,000                             & 10                                 & 0.056                       & 0.735                         & 2.346                        & \textbf{-0.001}                \\
100,000                            & 10                                 & 0.077                       & 1.519                         & 2.599                        & \textbf{-0.064}                \\ \hline
1,000                              & 20                                 & \textbf{0.055}              & 1.125                         & 4.012                        & \underline{0.109}                    \\
10,000                             & 20                                 & 0.061                       & 1.684                         & 4.990                        & \textbf{0.057}                 \\
100,000                            & 20                                 & 0.125                       & 3.626                         & 5.927                        & \textbf{0.003}                 \\ \hline
1,000                              & 50                                 & \textbf{0.098}              & 4.125                         & 14.84                        & \underline{0.594}                    \\
10,000                             & 50                                 & \textbf{0.120}              & 6.385                         & 19.65                        & \underline{0.548}                    \\
100,000                            & 50                                 & \textbf{0.298}              & 13.27                         & 22.37                        & \underline{0.507}                   
\end{tabular}
\caption{Time (s)} \label{fig_time}
\end{subtable}
\caption{Experimental results on synthetic data across a variety of metrics. AutoPC easily outperforms all other algorithms in accuracy across different metrics (a)-(c). AutoPC either comes in first or second place in terms of time (d); underlined values denote second place.} \label{fig_metrics}
\end{table}

\begin{table}
\centering
\smaller
\begin{tabular}{c|ccccc}
\rowcolor[HTML]{EFEFEF} 
\textit{Metric}                   & Mean & BIC            & StARS & OCTs  & AutoPC         \\ \hline
\cellcolor[HTML]{FFFFFF}SHD  & 5.238  & 5.272          & 5.096 & 5.316 & \textbf{4.888} \\
\cellcolor[HTML]{FFFFFF}F1   & 0.975  & 0.973          & 0.981 & 0.971 & \textbf{0.988} \\
\cellcolor[HTML]{FFFFFF}MCC  & 0.433  & 0.433          & 0.434 & 0.432 & \textbf{0.435} \\
\cellcolor[HTML]{FFFFFF}Time & Ref    & \textbf{0.072} & 4.877 & 1.974 & \underline{0.183}   
\end{tabular}
\caption{Real data results. AutoPC again achieves the best accuracy and the second fastest time.} \label{table_real}
\end{table}

\section{Conclusion} \label{sec_conclusion}
We proposed to select the $\alpha$ value of the PC algorithm by forcing it to double check its output with a second run. We called the resultant procedure AutoPC. We proved that AutoPC optimizes a user chosen metric in the oracle setting. The algorithm also selects a sequence $\widehat{\alpha}_n$ that asymptotically optimizes the metric in the sample limit. Experimental results highlighted the superiority of AutoPC in selecting the best $\alpha$ value across multiple evaluation criteria.

\section*{Acknowledgments}
TBD

\bibliographystyle{elsarticle-num-names}
\bibliography{biblio}

\end{document}